\documentclass[letterpaper, 10 pt, conference]{ieeeconf}  

\IEEEoverridecommandlockouts                              

\usepackage{graphics} 
\usepackage{epsfig} 
\usepackage{amsmath} 
\usepackage{amssymb}  
\usepackage{mathtools}
\usepackage{siunitx}
\usepackage{booktabs}
\usepackage{xcolor}
\usepackage[noadjust]{cite}
\usepackage{lipsum}
\usepackage{dblfloatfix}
\usepackage{booktabs}
\usepackage{multirow}
\usepackage{adjustbox}
\usepackage{hyperref}

\title{\LARGE \bf
A Modular 3-Degree-of-Freedom Force Sensor for \\ Robot-assisted Minimally Invasive Surgery Research
}

\author{Zonghe Chua$^{1}$ and Allison M. Okamura$^{2}$,~\IEEEmembership{Fellow, IEEE}
\thanks{*This work was supported by a Stanford Bio-X Fellowship and an National University of Singapore Development Grant.}
\thanks{$^{1}$Z. Chua is with the Electrical, Computer, and Systems Engineering Department, Case Western Reserve University, OH 44106, USA. {\tt\small zonghe.chua@case.edu}}%
\thanks{$^{2}$ A. M. Okamura is with the Mechanical Engineering Department, Stanford University, CA 94305, USA. {\tt\small aokamura@stanford.edu}}}%

\begin{document}

\bstctlcite{IEEEexample:BSTcontrol} 
\newcommand{\hlight}[1]{\textcolor{black}{#1}}

\maketitle
\thispagestyle{empty}
\pagestyle{empty}

\begin{abstract}

Effective force modulation during tissue manipulation is important for ensuring safe robot-assisted minimally invasive surgery (RMIS). Strict requirements for in-vivo distal force sensing have led to prior sensor designs that trade off ease of manufacture and integration against force measurement accuracy along the tool axis. These limitations have made collecting high-quality 3-degree-of-freedom (3-DoF) bimanual force data in RMIS inaccessible to researchers. We present a modular and manufacturable 3-DoF force sensor that integrates easily with an existing RMIS tool. We achieve this by relaxing biocompatibility and sterilizability requirements while utilizing commercial load cells and common electromechanical fabrication techniques. The sensor has a range of $\pm5\,\text{N}$ axially and $\pm 3\,\text{N}$ laterally with average root mean square errors (RMSEs) of below 0.15\,N in all directions. During teleoperated mock tissue manipulation tasks, a pair of jaw-mounted sensors achieved average RMSEs of below 0.15\,N in all directions. For grip force, it achieved an RMSE of 0.156\,N. The sensor has sufficient accuracy within the range of forces found in delicate manipulation tasks, with potential use in bimanual haptic feedback and robotic force control. As an open-source design, the sensors can be adapted to suit additional robotic applications outside of RMIS.
\end{abstract}



\section{Introduction}

Respect for tissue \cite{martin1997objective}, or force sensitivity \cite{goh2012global}, is considered an important skill for performing safe surgery and requires good control of applied forces. Thus, knowledge of the force exerted by a robotic system on the surgical environment is important during robot-assisted minimally invasive surgery (RMIS) to enable safe tissue handling. Force information can be used to provide haptic feedback to the surgeon, automatically and objectively evaluate their force sensitivity for training and credentialing purposes, and inform the decisions and movements of an autonomous agent. 

Force information has been difficult to obtain for the above purposes in RMIS because there is no native distal force sensing in commercial RMIS systems. This is due in part to designers needing to meet the strict requirements for biocompatibility and sterilizability of RMIS instruments while ensuring cost-effectiveness \cite{EnayatiReview2016,HadiHosseinabadi2022}. While researchers have explored many approaches to developing force sensors that attempt to address the above constraints, none have gained commercial adoption. Furthermore, many designs contain complex electromechanical components that require specialized knowledge to manufacture, assemble, or integrate. This limits their adoption even in the research community.

\begin{figure}[!t]
    \centering
    \vspace{1.8mm}
    \includegraphics[width=\linewidth]{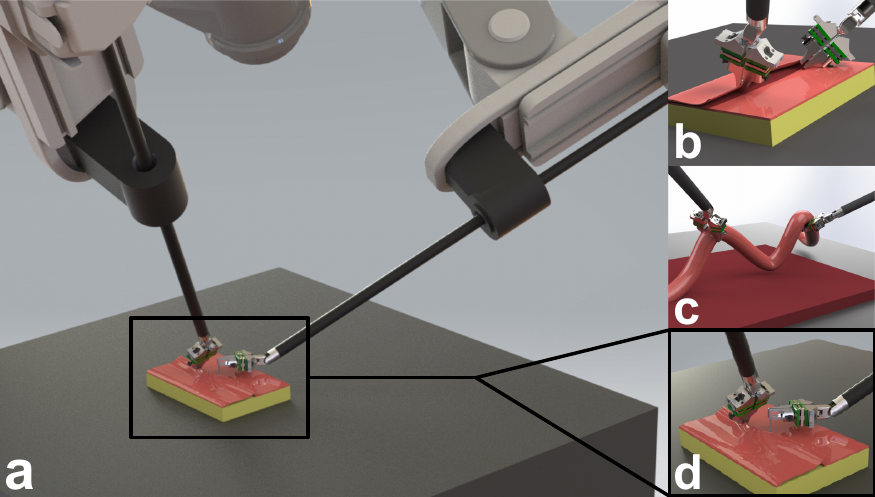}
    \caption{Concept renderings of the 3-DoF force sensor design in example use cases requiring bimanual manipulation. (a) View of sensorized patient-side manipulators during suturing. Close-up of sensorized forceps during (b) blunt dissection, (c) running the bowel, and (d) suturing.}
    \label{fig:concept}
\end{figure}

To make up for the lack of feasible force sensing options for RMIS tools, researchers aiming to improve force sensitivity have often relied on existing general-purpose commercially available force sensors like those from ATI Industrial Automation (Apex, NC, USA) that are placed in or under the artificial tissue being manipulated. In such a set up, researchers are limited to RMIS studies using only a single end effector \cite{Bahar2020,Chua20201Dynamics} or measuring a single force value for both end-effectors \cite{Brown2017apms,sensubGaleazzi2022}. This approach prevents the study of studying bimanual force-critical tasks such as those shown in Fig.\,\ref{fig:concept} and thus limits applicability to real surgery. 

 One approach that circumvents the need for end-effector force sensors is indirect force sensing. This has been explored using physics-based \cite{fontanelli2017modelling,Wang2019dynamic} or neural network models \cite{Yilmaz2020dynamic} of the robot to predict joint torques and using vision-based finite element \cite{Haouchine2018FEA} or deep learning methods \cite{aviles2014recurrent,aviles2016towards,Marban2019rnn,ChuaNeuralForce}. However, these approaches need to be trained or benchmarked against a ground truth. To achieve this, researchers have often used a single environmental force sensor like the ATI sensors noted above. For methods that rely entirely on the robot's internal state, this approach is feasible because each end effector can be trained separately. However, for methods that rely on measuring environmental changes, such as vision-based methods, this approach has limited applicability to bimanual manipulations where internal ground truth force data cannot easily be resolved.

In this work, we present the the design and characterization of a 3-degree-of-freedom (3-DoF) force sensor for RMIS research. By relaxing the strict constraints on size, biocompatibility, and sterilizability, we realize a sensor that is easy to manufacture and integrate into existing hardware while achieving satisfactory performance for a wide variety of research use cases. To capitalize on its manufacturability and modularity, we have open-sourced the design to enable researchers to adapt the sensor for their desired application both within RMIS and for other robotic applications where deploying commercial sensors is difficult due to size, cost, and customizability requirements.

\section{Background}





Previous works have placed sensors at different locations of the RMIS instrument. These include the jaws, wrist, lower shaft, trocar, upper shaft, and the instrument base. These works have also employed various types of sensing technology, with metal strain gauges, capacitive sensors, fiber-Bragg gratings, and infrared (IR) light intensity measurement being the most common technologies employed.

Force sensors located at the jaw have been implemented as both 2-DoF \cite{Hong2012} and 3-DoF sensors \cite{Yu2018} using strain gauges and custom jaw flexures. Custom jaw flexures were also employed in \cite{Kim2018} for 3-DoF force and 2-DoF moment sensing using capacitive elements. Such jaw flexures allow forces to be sensed when any part of the tool tip interacts with the environment. This is important during tasks such as blunt dissection (Fig.\,\ref{fig:concept}b) and running the bowel (Fig.\ref{fig:concept}c) where the tip and back of the tool tip are used for manipulation. This is in contrast to approaches which place sensors on the grasping surface of each jaw and thus only allow forces to be sensed when the environment is grasped \cite{Kim2015,Dai2017}. By locating the sensors at the jaw, grip force can also be computed from the force measurements at each jaw.

Locating force sensors above the articulated tool wrist reduces the electromechanical integration complexity typical of sensors located at the jaw. In \cite{Li2017wrist}, a Stewart platform with strain gauges was used to measure 3-DoF forces at the articulated wrist of a custom RMIS tool. In \cite{Lee2015}, 6-DoF forces and moments were measured at the articulated wrist using capacitive sensors. Torque sensors were also embedded in the drive pulleys and used to both measure grip force and compensate for noise in the wrist force sensors due to drive cable actuation. Sensors have also been located on the lower shaft of RMIS tools with both \cite{Shahzada2016fbgshaft} and \cite{Du2022fbgshaft} using fiber-Bragg gratings to measure 2-DoF lateral forces.

Sensors have also been placed at the interface of the patient's body (at the trocar), or outside the body, for example, on the upper shaft or instrument base. In \cite{Kim2017trocar}, strain gauges were used at the trocar to measure 2-DoF lateral forces, while in \cite{Fontanelli2020trocar}, IR intensity measurement was used instead. At the upper shaft, \cite{hadi2021shaft} used IR intensity to measure 3-DoF forces and moments though for the three directions of force, accuracy metrics for only the two lateral directions of force were reported. At the instrument base and upper shaft, \cite{Novoseltseva2018} used strain gauges to measure 3-DoF forces, with force measurements along the main axis of tool showing poorer accuracy relative to those along the lateral directions.

\section{Methods}

\subsection{Target Design Requirements}

To realize a force sensor that is accessible to the research community, the design of the sensor should be easily manufacturable and integrated into existing RMIS tools. This makes sensors located on the upper shaft of the instrument base particularly suitable \cite{HadiHosseinabadi2022}. However, these designs typically lack accuracy along the main axis of the RMIS tool. Furthermore, they are unable to measure grip force, which can be useful for evaluating surgical skill or for providing feedback to improve tissue manipulation. 

RMIS research is often performed on ex-vivo or dry lab tasks. This relaxes the requirements on biocompatibility, sterilizability, and size. Thus, placing sensors at the tool jaws does not require complex jaw designs and can utilize small-size commercial load cells. At the same time, locating the sensor at the jaw reduces measurement noise and enables more accurate measurement of force along the main axis of the tool. Additionally, the jaw sensor placement allows for straightforward grip force measurement. 

Based on these considerations, we designed a jaw-mounted sensor that can be customized to suit different RMIS tools and different use cases. Our target use case of tissue manipulation requires that all parts of the jaw be able to sense force, and thus, unlike in \cite{Kim2015,Dai2017}, the sensing elements cannot be solely mounted on the grasping surfaces of each jaw.

Tissue manipulation forces can be up to 3.8\,N in the lateral direction, $-$10.3\,N in the axial compression direction, and 7.8\,N in the axial retraction directions \cite{Toledo1999}. However, there is a need to balance these requirements against the current capabilities of small size commercial load cells. Based on these considerations, our sensor target force ranges are $\pm3\,\text{N}$ in the lateral directions and $\pm5\,\text{N}$ in the axial direction. In terms of accuracy requirements, the average kinesthetic force just noticeable difference for the human hand is 12.5\% \cite{Vicentini2010}, and thus the sensor requires a minimum sensor accuracy of 0.375\,N in all directions for error imperceptibility.


\subsection{Electromechanical Design}
\begin{figure}[!t]
    \centering
    \vspace{1.8mm}
    \includegraphics[width=0.75\linewidth]{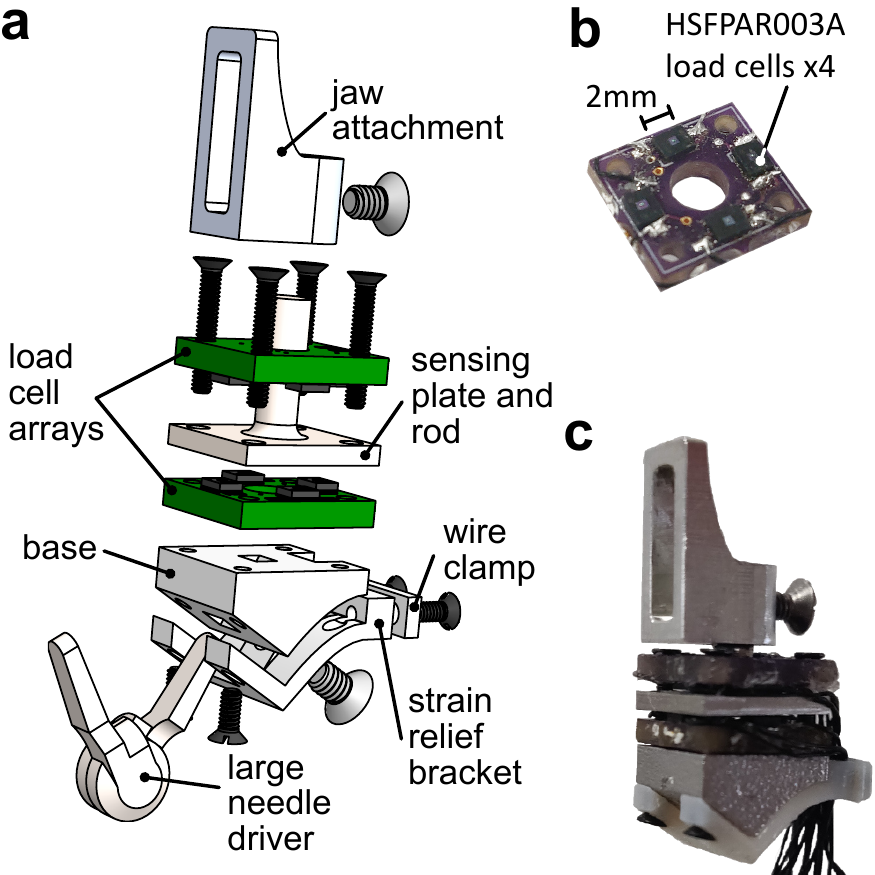}
    \caption[Electromechanical design of the force sensor.]{(a) Exploded view of the force sensor mounted to one jaw of the da Vinci large needle driver. (b) Arrangement of the load cells on the PCB of the load cell array. (c) Fully assembled force sensor.}
    \label{fig:sensor_mech}
\end{figure}

Based on the above design requirements, we designed a 3-degree-of-freedom force sensor located at the tool jaws. As shown in Fig.\,\ref{fig:sensor_mech}a, the sensor comprises five main parts: (1) the base, (2) two load cell arrays, (3) the sensing plate and rod, and (4) the jaw attachment. An optional strain relief bracket (also shown in Fig.\,\ref{fig:sensor_mech}a) can be added to help secure and route wires.

The base is 3D-printed in 6061 aluminum and its geometry can be modified to interface with different RMIS tool jaws. In this paper we present a design that interfaces with the da Vinci Surgical System (Intuitive Surgical, Inc., Sunnyvale, CA, USA) large needle driver jaw using a M2$\times$3 set screw. One of two load cell arrays is placed above the top surface of the base and is electrically isolated using Kapton tape. 

\begin{figure}[!t]
    \centering
    \includegraphics[width=0.75\linewidth]{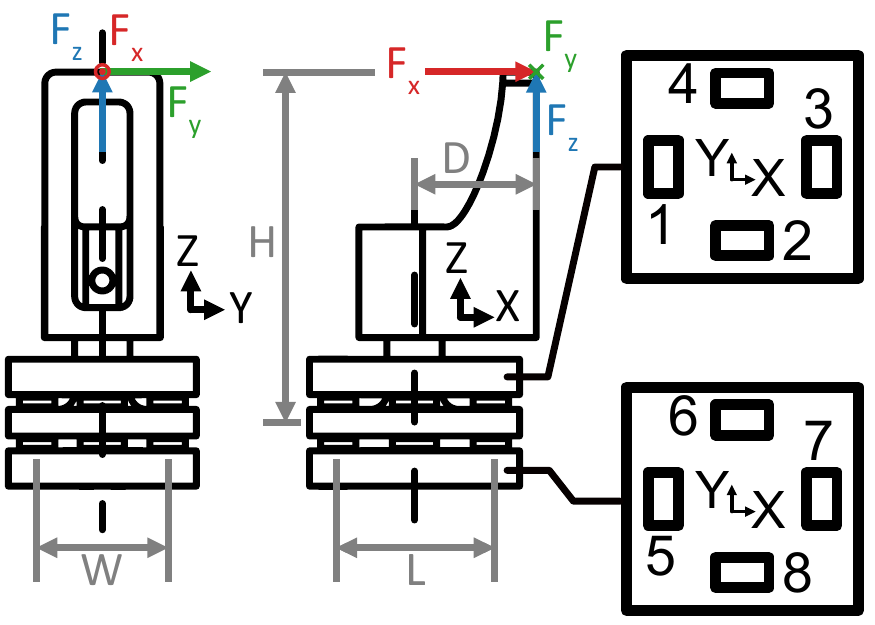}
    \caption[Front and side views of the force sensor with top views of the load cell arrays.]{Front and side views of the force sensor with top views of the load cell arrays. Load cells are numbered 1 through 8 corresponding to Eqns.\,(\ref{eqn:moment_x})\,--\,(\ref{eqn:force_z}).}
    \label{fig:FBD}
\end{figure}

Each load cell array is a 9.5$\times$8.5$\times$1.6\,mm 2-layer FR4 printed circuit board with four HSFPAR003A load cells (Alps Alpine, Japan) soldered along the perimeter (Fig.\,\ref{fig:sensor_mech}b). The load cells measure compression forces of up to 8\,N and rely on a piezoresistive full Wheatstone bridge which allows for good temperature stability. The bridge outputs are amplified using AD623 (Analog Devices, Norwood, MA, USA) instrumentation amplifiers with a gain of 21. This results in a sensor response of 3.063\,$\text{N}\,\text{V}^{-1}$. The amplified analog signals from each sensor were recorded on a PC using an Arduino Mega with serial communication at 125\,Hz.

The sensing plate and rod is machined out of 303 stainless steel for high stiffness. A second load cell array is placed on the top face of the plate in opposition to the first load cell array. The two load cell arrays and the sensing rod and plate are attached to the base using four M1.2$\times$12\,mm screws. 

A jaw attachment, which replaces the original tool jaws for grasping, is machined out of 6061 aluminum and is attached to the sensing rod by a M2$\times$3\,mm set screw. The jaw attachment is interchangeable, allowing for researchers to machine different shapes to suit the task they are studying. Here we fabricated a generic shape for tissue retraction and palpation that has a height of 12\,mm. During manufacturing of the load cell arrays, there are small deviations in the heights of each load cell after soldering. Thus, we enabled consistent contact between the sensing plate and the individual load cells on each side by inserting metal shims. The sensors were preloaded up to a maximum of 1.5\,N. The fully assembled sensor is shown in Fig.\,\ref{fig:sensor_mech}c with overall dimensions of 9.5$\times$8.5$\times$23.8mm and a weight of 3.33\,g.

\subsection{Sensing Principle}
\label{sensingprinciple}
The sensing principle relies on the moment balance about the lateral axes of the device, henceforth referred to as the sensor x- and y-axis, and force balance in the main axis, henceforth referred to as the sensor z-axis. Assuming an interaction at the tip of the jaw attachment, and neglecting the contribution of shear forces to the moment balance, the resulting force and moment equations are

\begin{equation}
    M_x =  F_y H - \frac{Lc}{2} (v_2 + v_6- v_4 - v_8) \text{,}
\label{eqn:moment_x}
\end{equation}
\begin{equation}
    M_y =  F_x H - F_z D - \frac{Wc}{2} (v_1 + v_7- v_3 - v_5) \text{, and}
\label{eqn:moment_y}
\end{equation}
\begin{equation}
    F_z =  c (v_1+v_2+v_3+v_4-v_5-v_6-v_7-v_8)\text{,}
\label{eqn:force_z}
\end{equation}
where $H=15.85$\,mm, $D=5.50$\,mm,  $L=3.45$\,mm, and  $W=2.95$\,mm. $c=3.063\,\text{N}\,\text{V}^{-1}$ is the voltage change per unit force, and $v_i$ the voltage output of the $i\textsuperscript{th}$ sensor as labeled in Fig.\,\ref{fig:FBD}. From Eqns. (\ref{eqn:moment_x})\,--\,(\ref{eqn:force_z}), we can express the measured forces as
\begin{equation}
    \begin{bmatrix}
    F_x \\ F_y \\ F_z
    \end{bmatrix}
    = A
    \begin{bmatrix}
    v_1 \\ \vdots \\ v_8
    \end{bmatrix}\text{,}
\label{eqn:force_voltage}
\end{equation}
where $A\in	\mathbb{R}^{8\times3}$ is a sensitivity matrix that maps sensor voltage outputs to forces. 


\subsection{Calibration Method}
The  actual value of $A$ can be estimated through linear least-squares fitting to calibration data. To improve the quality of calibration, we also fit a constant offset term for each direction of force. This results in the estimated sensitivity matrix $A^+$ having a dimension of $3\times9$.

\begin{figure}
    \centering
    \vspace{1.8mm}
    \includegraphics[width=0.6\linewidth]{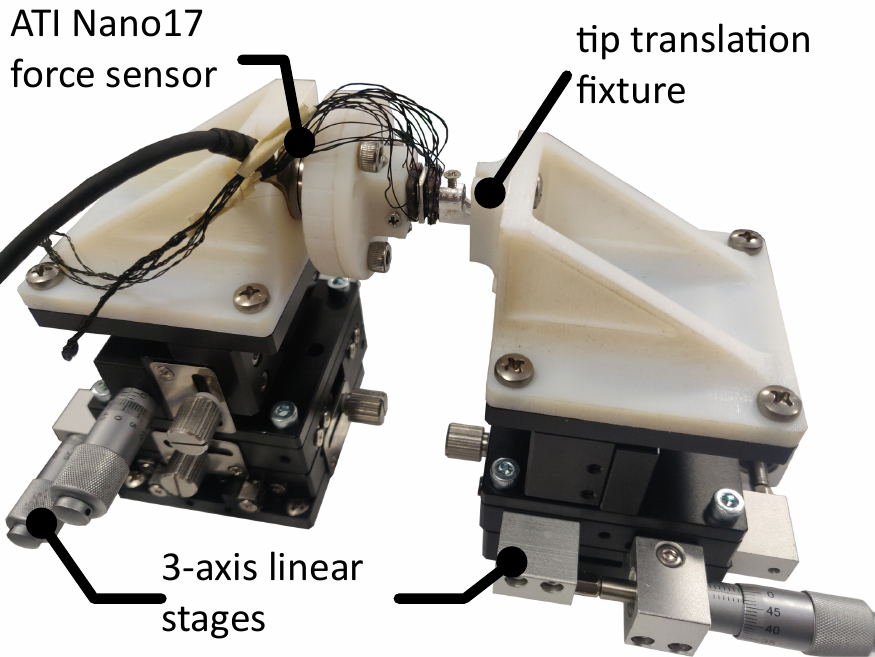}
    \caption{Setup used for static calibration of a single force sensor.}
    \label{fig:static_cal}
\end{figure}

To perform the calibration, the sensor was mounted on a Nano17 force sensor (ATI, Apex, NC, USA). The tip of the jaw attachment was affixed to a 3-axis linear stage as shown in Fig.\,\ref{fig:static_cal} and loaded in each Cartesian axis in increments of $0.5\pm0.1\,\,\text{N}$ through the target sensing range of 0 to $\pm3$\,N in the x- and y-directions and 0 to $\pm5$\,N in the z-direction. To reduce calibration errors due to possible hysteretic behavior, data was collected during loading and unloading. The quality of the calibration was evaluated using the root mean square error (RMSE), the normalized root mean square deviation (NRMSD) which is the RMSE normalized by the measurement range of the sensor, the coefficient of determination ($R^2$), and the hysteresis, which is the maximum difference between corresponding measured forces during loading and unloading normalized by the maximum force \cite{figliola2020hysteresis}.

\begin{figure}[!t]
    \centering
    \includegraphics[width=\linewidth]{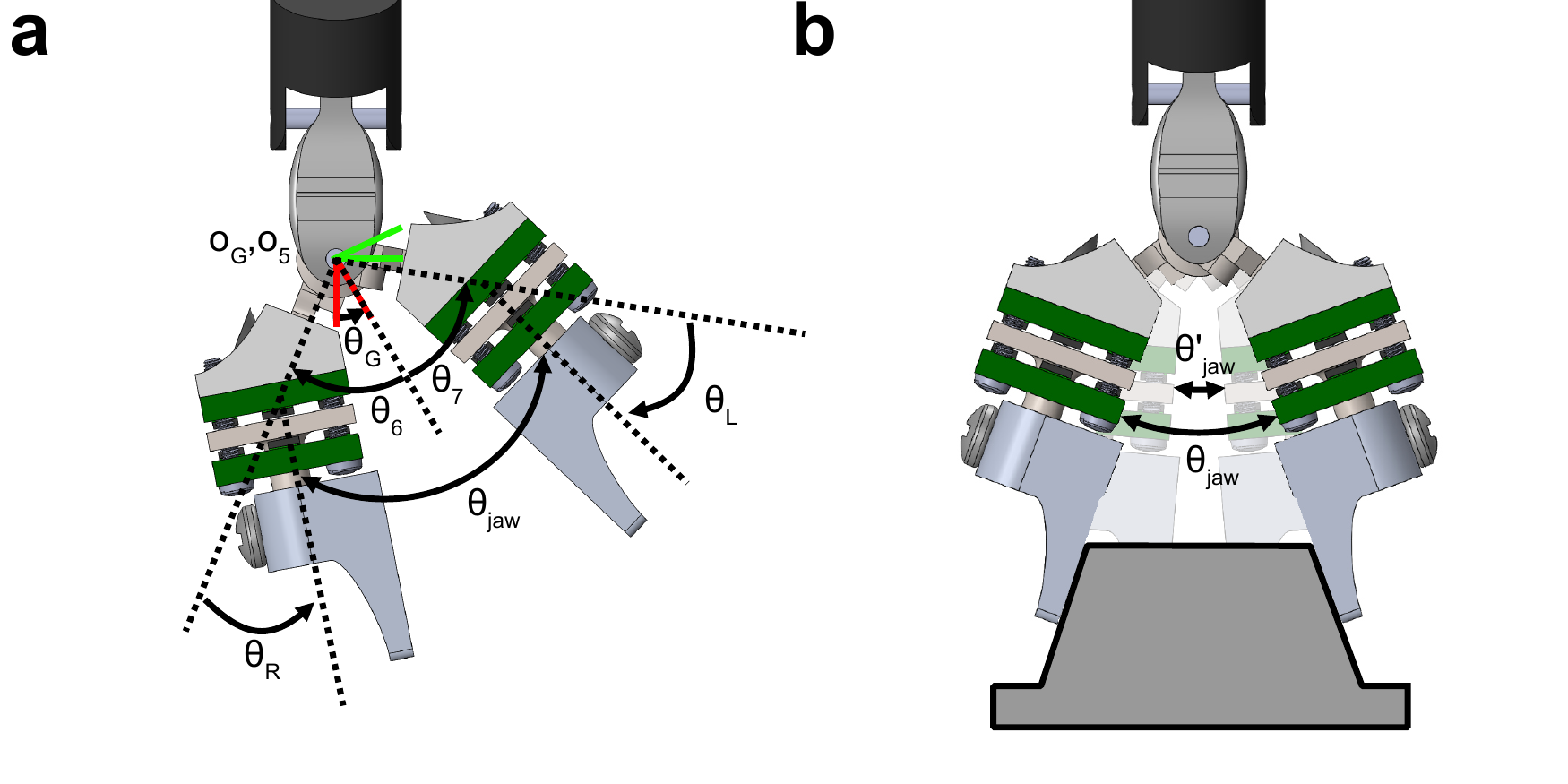}
    \caption[Gripper joint angle definitions.]{(a) Definitions of angles at the gripper of the sensorized RMIS tool. (b) Angle definitions for the software reported jaw angle $\theta^{'}_{\text{jaw}}$ and the actual jaw angle $\theta_{\text{jaw}}$.}
    \label{fig:gripper_angles}
\end{figure}

\begin{figure*}[!t]
\centering
\vspace{1.8mm}
    \includegraphics[width=\linewidth]{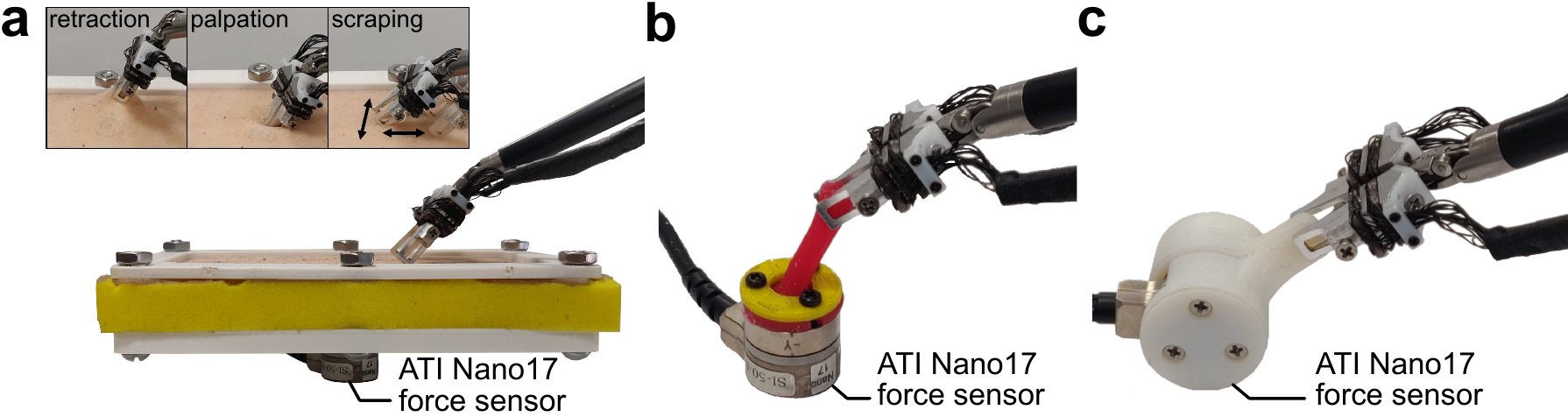}
    \caption{Setups for evaluating performance of force sensors (shown here with strain relief brackets) when mounted on the da Vinci Large Needle Driver tool. (a) The flat tissue manipulation (with inset showing from left to right, retraction, palpation and scraping movements), (b) the cylindrical stem manipulation, and (c) grip force measurement setups.}
    \label{fig:dual_jaw_exp}
\end{figure*}

\subsection{Performance Evaluation}

\subsubsection{Single Jaw Evaluation}
\label{ch4:singlejawevalmethod}
The single jaw evaluation was performed by exerting varying loads on the tip of the jaw attachment in all three Cartesian directions while the sensor was mounted to a Nano17 force sensor using the calibration setup shown in Fig.\,\ref{fig:static_cal} without the tip translation fixture. The sensor accuracy was determined by computing the RMSE and NRMSD of the force measurements and averaging them over three trials. The maximum error over three trials was also calculated. 

\subsubsection{Dual Jaw Evaluation}
\label{dualjawevalmethod}
To measure the manipulation forces at the end-effector, each jaw of the da Vinci large needle driver needs to be instrumented with a force sensor. The forces measured in the reference frame of each sensor are first resolved into the reference frame of a da Vinci Research Kit (dVRK) \cite{kazanzides2014dvrk}. This is done using the robot forward kinematic model and the joint position estimates from the motor encoders (6 joints and the gripper angle) to obtain the individual jaw poses in the robot reference frame. The resultant force, $F_r$, is thus
\begin{multline}
    \prescript{0}{}{F_r} =  \prescript{0}{6}{T}(\theta_1,\dots,\theta_6,\theta_G) \prescript{6}{R}{T}(\theta_R) \prescript{R}{}{F} \\ + \prescript{0}{7}{T}(\theta_1,\dots,\theta_5,\theta_7,\theta_G) \prescript{7}{L}{T}(\theta_L) \prescript{L}{}{F} 
\label{eqn:resultantforce}
\end{multline}
where $\prescript{j}{i}{T}$ are transformation matrices describing transformations mapping frame $i$ to $j$, with frame 0 being the dVRK origin, frames 6 and 7 describing the left and right tool gripper jaws respectively, and the L and R frames describing the local coordinate frames of each force sensor. The values of $\theta_{0}$ to $\theta_7$ are joint rotation angles, $\theta_R$ and $\theta_L$ are fixed rotation angles, and $\theta_G$ is the angle between the x-axis of frame 5 and the bisector of $\theta_6$ and $\theta_7$ (Fig.\,\ref{fig:gripper_angles}a).

Due to backlash and stretching of the tool actuation tendons, the computed jaw angle $\theta^{'}_\text{jaw}$ during grasping, as derived from the estimates of $\theta_6$ and $\theta_7$ and values of $\theta_R$ and $\theta_L$, is smaller than the actual sensorized tool jaw angle $\theta_\text{jaw}$ (Fig.\,\ref{fig:gripper_angles}b). To ensure that the correct joint angles are used during the pose computation, we define $\theta_\text{jaw}$ such that
\begin{equation}
    \theta_\text{jaw} = (\theta_6 - \theta_R) + (\theta_7 - \theta_L) = 
\begin{cases}
    \theta^{'}_\text{jaw} ,& \text{if } \theta^{'}_\text{jaw}>\theta_\text{min}\\
    \theta_\text{min},              & \text{otherwise}
\end{cases}
\label{eqn:gripper jaw}
\end{equation}
where $\theta_6=\theta_7$, and $\theta_\text{min}$ is the minimum jaw angle during grasp. 

Because each jaw is instrumented with a force sensor, the grasp force between the two jaws can be obtained. The grasp force was computed by using a two-point grasp model and resolving the forces measured at each sensor into the line of action between the two grasp points. Applying the rules derived by Yoshikawa and Nagai \cite{Yoshikawa1991}, the grasp force for a two-point grasp is \hlight{
\begin{equation}
    F_g = \text{min}(\, |(\prescript{G}{R}{T}\prescript{R}{}{F})\cdot\hat{j}| \, , \, |(\prescript{G}{L}{T}\prescript{L}{}{F})\cdot\hat{j}|)\text{,}
    \label{eqn:grasp}
\end{equation}
where $G$ is denotes the gripper frame of reference as shown in Fig.\,\ref{fig:gripper_angles}a and $\hat{j}=\begin{bmatrix}0&1&0\end{bmatrix}^\intercal$.}

To evaluate the sensor on realistic tissue manipulation tasks, we designed two environments that enable different types of manipulation forces to be exerted by an instrumented RMIS tool mounted on a teleoperated dVRK. The first environment consisted of an artificial silicone tissue (Limbs and Things, Savannah, GA, USA) placed over a sponge with a Nano17 force sensor placed underneath (Fig.\,\ref{fig:dual_jaw_exp}a). In this environment the tool can be teleoperated to perform palpation, scraping, and tissue retraction. However, due to the low friction of the silicone as well as the need to limit grasp forces in software to protect the sensor from damage, the tissue retraction force achievable in this environment were low compared to the sensor's operating range. The second environment consisted of a cylindrical silicone stem mounted on top of a Nano17 force sensor (Fig.\,\ref{fig:dual_jaw_exp}b). This setup thus allowed the teleoperator to exert higher retraction forces on the environment. In these two setups, the ground truth force during the teleoperated interactions can be obtained and used to evaluate the RMSE, RMSD, and maximum error of the resultant force measured from the dual jaw sensors over three trials of each task. However, the grip forces cannot be evaluated.

\begin{table*}[!t]
\centering
\caption{Summary of static calibration results for both sensors.}
\begin{tabular}{ccccccccccccc} 
\toprule
\multirow{2}{*}{Sensor} & \multicolumn{3}{c}{RMSE (N)} & \multicolumn{3}{c}{NRMSD (\%)} & \multicolumn{3}{c}{$R^2$} & \multicolumn{3}{c}{Hysteresis (\%)}  \\
                        & x     & y     & z            & x     & y     & z              & x     & y     & z      & x+, x-    & x+, x-    & z+, z-       \\ 
\cmidrule(r){1-1}\cmidrule(lr){2-4}\cmidrule(lr){5-7}\cmidrule(lr){8-10}\cmidrule(r){11-13}
A                       & 0.023 & 0.056 & 0.044        & 0.388 & 0.876 & 0.438          & 0.999 & 0.996 & 0.999  & 3.96, 3.48 & 3.25, 2.85 & 2.78, 2.37    \\
B                       & 0.032 & 0.048 & 0.045        & 0.531 & 0.814 & 0.445          & 0.999 & 0.997 & 0.999  & 2.13, 2.73 & 2.15, 1.62 & 3.17, 3.36    \\
\bottomrule
\end{tabular}
\label{tbl:static_cal}
\end{table*}

To evaluate grip force, a separate experiment was devised. This involved attaching 3D-printed cantilevers to each interface of the Nano17 force sensor, leaving the sensor ungrounded as shown in Fig.\,\ref{fig:dual_jaw_exp}c. The sensorized RMIS tool was then teleoperated to grasp and release the opposing cantilevers five times each in three different trials. The sensor accuracy was determined by computing the RMSE and normalized root mean square deviation of the force measurements and averaging them over three trials. The maximum error over three trials was also calculated.

\section{Results and Discussion}

\subsection{Static Calibration}

\begin{figure}[!t]
    \centering
    \includegraphics[width=\linewidth]{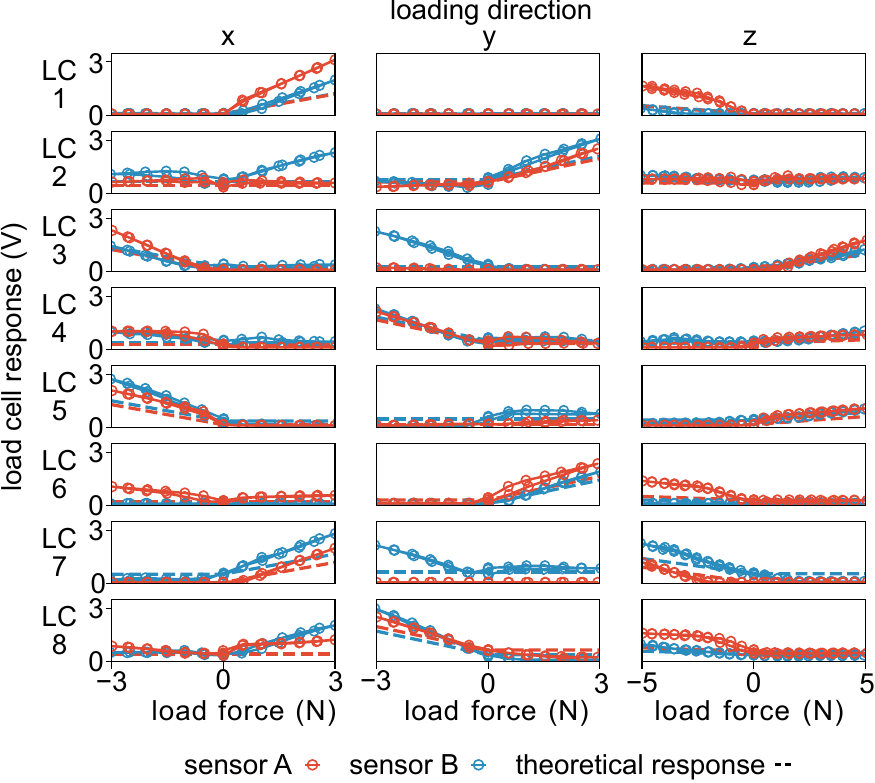}
    \caption{Load cell (LC) response under loading in each Cartesian direction. Load forces were measured using an ATI Nano17 force sensor within the calibration set up shown in Fig.\,\ref{fig:static_cal}.}
    \label{fig:sensor_response}
\end{figure}

\begin{figure}[!t]
    \centering
    \includegraphics[width=\linewidth]{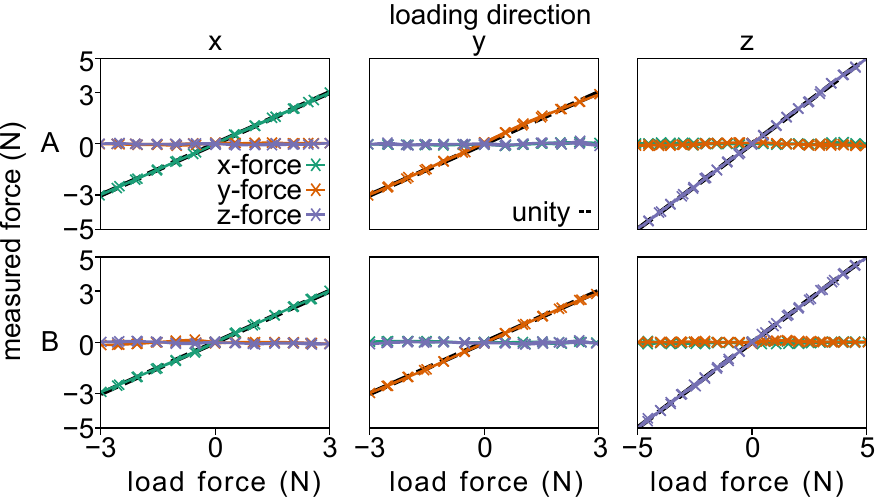}
    \caption{Forces measured by the force sensors in each axis when loaded and unloaded independently in each Cartesian direction.}
    \label{fig:calibration_response}
\end{figure}

\begin{table*}[!t]
\centering
\vspace{1.8mm}
\caption{Summary of single jaw evaluation results for both sensors.}
\begin{tabular}{cccccccccc} 
\toprule
\multirow{2}{*}{Sensor} & \multicolumn{3}{c}{RMSE (N)} & \multicolumn{3}{c}{NRMSD (\%)} & \multicolumn{3}{c}{Max Error (N)}  \\
                        & x     & y     & z            & x     & y     & z              & x     & y     & z                  \\ 
\cmidrule(lr){1-1}\cmidrule(lr){2-4}\cmidrule(lr){5-7}\cmidrule(lr){8-10}
A                       & 0.111$\pm$0.016& 0.105$\pm$0.015& 0.064$\pm$0.004       & 1.845$\pm$0.278 & 1.745$\pm$0.246 & 0.635$\pm$0.036          & 0.483 & 0.415 & 0.325              \\
B                       & 0.117$\pm$019& 0.146$\pm$0.013& 0.126$\pm$0.012        & 1.945$\pm$0.316 & 2.43$\pm$0.220  & 1.264$\pm$0.123          & 0.573 & 0.654 & 0.536              \\
\bottomrule
\end{tabular}
\label{table:single_jaw}
\end{table*}



\hlight{As described in Sec. \ref{dualjawevalmethod}, sensing of manipulation and grip forces at the end-effector required two sensors to be fabricated. Thus, the calibration was performed on two sensors, A and B, each corresponding to one jaw. The load cell responses during loading in each Cartesian direction during calibration are shown in Fig.\,\ref{fig:sensor_response} and indicate that when the sensing principle described in Sec. \ref{sensingprinciple} predicted a response  (dashed lines) from a given load cell, there was an appropriate response from that corresponding load cell. Additionally, there were some unexpected responses when no response was predicted due to sensor crosstalk and uneven plate contact that arose from slight errors in manufacturing. Overall, the sensors displayed good linearity over its functional range (Fig.\,\ref{fig:calibration_response}), with the redundant sensing architecture of the sensor mitigating any detrimental effects of crosstalk on sensor performance. The low deviation from unity in both loading and unloading seen in Fig.\,\ref{fig:calibration_response} also indicates that the sensor has low hysteresis despite its non-monolithic design. The results of the calibration procedure are summarized Table \ref{tbl:static_cal}. }



\subsection{Single Jaw Evaluation}

\begin{figure}[!t]
    \centering
    \includegraphics[width=0.9\linewidth]{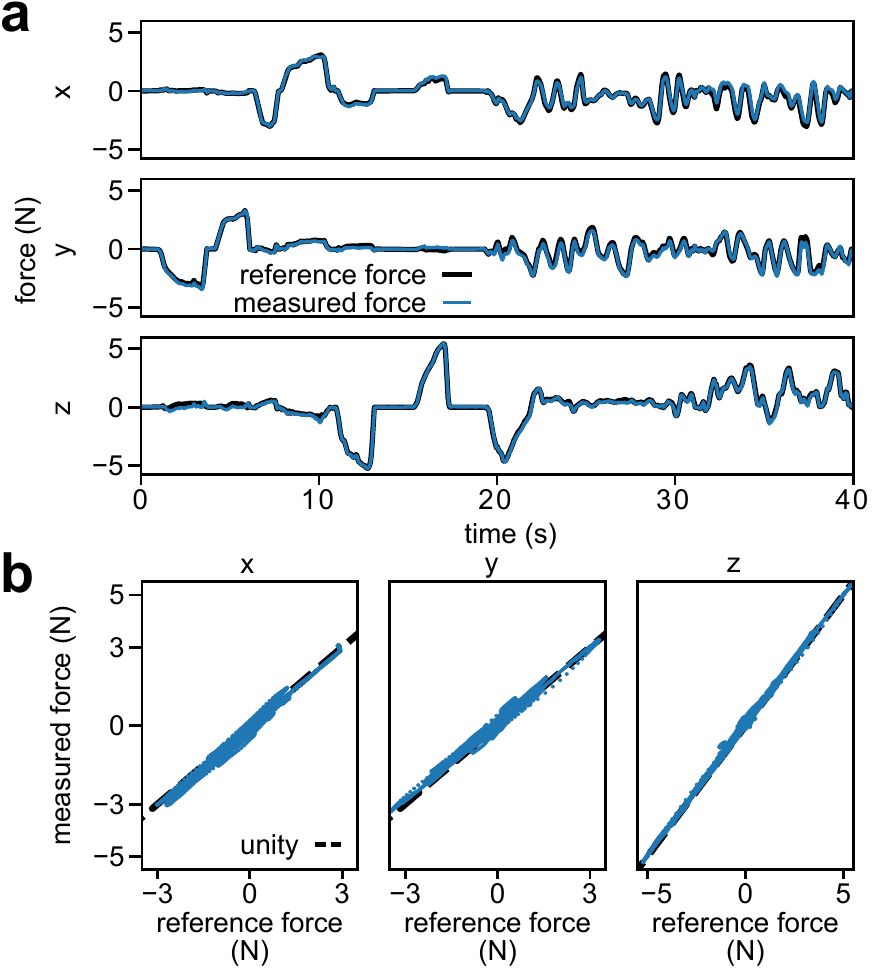}
    \caption{Selected results from single jaw evaluation of sensor B. (a) Recorded forces over time. (b) Measured force from the force sensor versus the reference force in each of the three Cartesian directions.}
    \label{fig:single_jaw_eval}
\end{figure}

The results of the single jaw evaluation are summarized in Table \ref{table:single_jaw}. The low RMSEs (up to 0.146\,N for sensor B in the y-direction) showed that the sensor performance seen in calibration translated to the good performance in the dynamic loading scenario of the single jaw evaluation. As seen in Fig\,\ref{fig:single_jaw_eval}a, from 25\,s to 40\,s, the sensor could accurately track fast changes in applied force while maintaining the desired accuracy of below 0.375\,N RMSE. The sensor showed high maximum error (up to 0.654\,N for sensor B in the y direction) when loaded in multiple directions with one direction loaded close to max of the calibration range. Thus, the measured versus reference force plot in Fig.\,\ref{fig:single_jaw_eval}b, show some deviation from unity for all three Cartesian directions. 

\subsection{Dual Jaw Evaluation}

\begin{figure}[!t]
    \centering
    \includegraphics[width=0.9\linewidth]{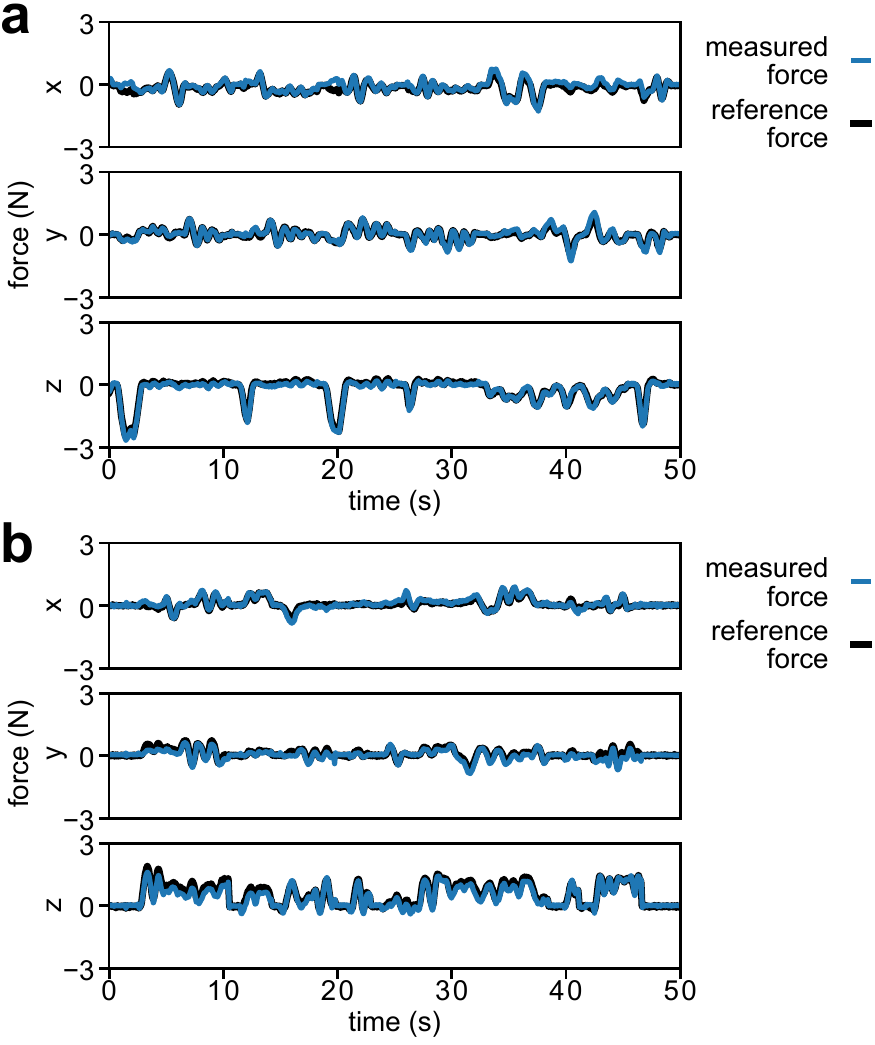}
    \caption[Selected results for the dual jaw evaluation tasks.]{Selected results for the (a) flat tissue  and (b) cylindrical stem manipulation tasks.}
    \label{fig:dual_jaw_eval}
\end{figure}

For both of the dual jaw evaluation tasks, we used a minimum jaw angle of $\theta_\text{min}=8.4^\circ$. This was required because the gripper does not fully close during grasping and thus the dVRK would report an incorrect gripper pose. This minimum jaw angle was empirically determined to reduce the error in the x-direction force measurements of both tasks with the RMSE of the sensor being below the minimum target of 0.375\,N as summarized in Table \ref{tbl:dual_jaw}. The force measurements from the sensor with respect to the ground truth for selected trials of the flat tissue and cylindrical stem manipulation tasks are shown in Fig.\,\ref{fig:dual_jaw_eval}.\hlight{ The plots show good tracking performance for lateral forces (x and y-direction in Fig.\,\ref{fig:dual_jaw_eval}a and b), as well as for palpation (z-direction in Fig.\,\ref{fig:dual_jaw_eval}a), and tension (z-direction in Fig.\,\ref{fig:dual_jaw_eval}b z-direction). }

\hlight{In our evaluation we identified three main sources of error. First, is the pose uncertainty of the jaws during grasping due to the cable-driven design of the dVRK robot where encoders are not placed directly on each joint of the surgical tool. Second, is the varying point of force application on the sensor, in which its calibration was done with forces applied only to the tip of the jaw attachment (Fig.\,\ref{fig:FBD} and Fig.\,\ref{fig:static_cal}). Third, is the small misalignment between the robot base coordinate frame and the reference force sensor coordinate frame.} 

\begin{table*}[!t]
\centering
\vspace{1.8mm}
\caption{Summary of dual jaw evaluation results for the flat tissue and cylindrical stem manipulation tasks.}
\begin{tabular}{cccccccccc} 
\toprule
\multirow{2}{*}{Task} & \multicolumn{3}{c}{RMSE (N)} & \multicolumn{3}{c}{NRMSD (\%)} & \multicolumn{3}{c}{Max Error (N)}  \\
                      & x & y & z                    & x     & y     & z              & x     & y     & z                  \\ 
\cmidrule(lr){1-1}\cmidrule(lr){2-4}\cmidrule(r){5-7}\cmidrule(lr){8-10}
Flat tissue &0.142$\pm$0.020& 0.078$\pm$0.013& 0.097$\pm$0.008 & 2.367$\pm$0.338 & 1.300$\pm$0.225 & 0.980$\pm$0.082          & 0.725 & 0.452 & 0.458              \\
Cylindrical stem &0.089$\pm$0.020& 0.149$\pm$0.023& 0.139$\pm$0.012 & 1.485$\pm$0.327 & 2.481$\pm$0.389 & 1.392$\pm$0.117          & 0.357 & 0.484 & 0.458              \\
\bottomrule
\end{tabular}
\label{tbl:dual_jaw}
\end{table*}

\subsection{Grip Force Evaluation}
\label{gripforceeval}

\begin{figure}[!t]
    \centering
    \includegraphics[width=0.9\linewidth]{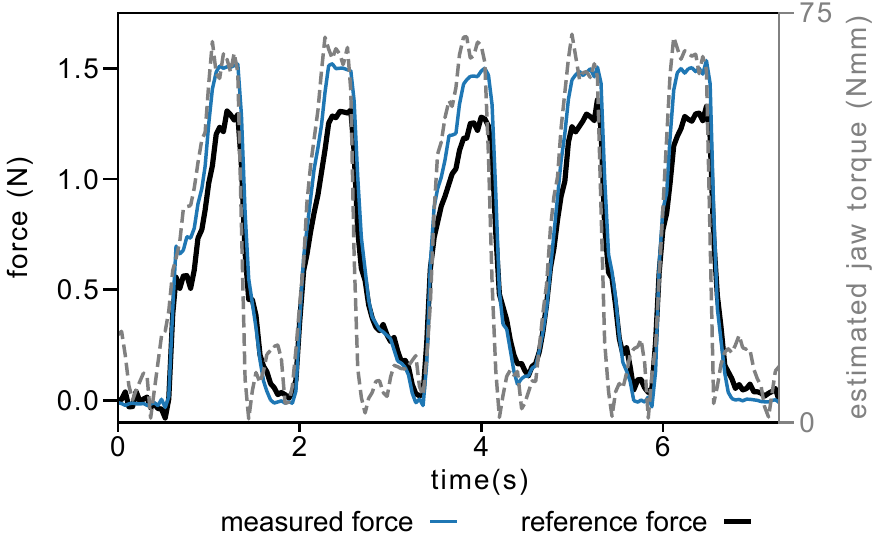}
    \caption[Selected grip force measurement result.]{Selected grip force measurement result. Estimated jaw torque is derived from the motor current measurements at the instrument base as reported through the dVRK API.}
    \label{fig:grip_eval}
\end{figure}

For the grip force evaluation, the RMSEs and standard deviations over three trials were 0.156$\pm$0.017\,N, and the maximum errors were 0.287\,N. A sample force plot over time using the default minimum jaw angle is shown in Fig.\,\ref{fig:grip_eval}. 

The dVRK requires the teleoperator to momentarily close the jaws to trigger teleoperation, this movement causes the tool jaws to snap together, resulting in an impulsive load on the sensors. To prevent damage to the sensors during this movement, we limited the maximum grip force in software. Because of this limit, the highest peak grip force achieved during the evaluation was 1.45\,N and the average peak grip force was 1.35\,N which is below the range of our sensor. \hlight{The limit also prevented us from evaluating the dual jaw performance of the sensor in both manipulation tasks up to the same range used in single jaw evaluation. The sample would slip from grasp before the higher forces were reached. With modification of the underlying dVRK teleoperation code, it would be possible for the sensors to be tested to its lateral force upper range of 3\,N.}

 
\section{Conclusion}

In this work, we presented a 3-DoF force sensor specifically designed to facilitate RMIS research. The sensor can be used with an existing RMIS tool, is manufacturable using low-volume manufacturing techniques, and has interchangeable jaws that allow for adaptation to different RMIS tasks. 

The current design is tolerant to manufacturing variability, leading to robust performance despite their presence. In both single-jaw standalone evaluation and dual-jaw evaluation on an RMIS tool, the sensor met the target accuracy specification of less than 0.375\,N RMSE. In single jaw evaluation, this accuracy was verified within the target sensing range of $\pm3$\,N for the lateral (x and y) directions and $\pm5$ for the axial (z) direction of force. Future work will investigate approaches to increase its robustness to manufacturing variability and enhance the assembly rigidity while maintaining or enhancing the sensing range.

In the dual jaw evaluation, the chief contributor to error was the pose uncertainty of the tool wrist and jaws. The uncertainty was most pronounced during high force manipulations and in grasping. To enable consistent, maximally accurate force measurements in cable-driven RMIS platforms like the dVRK, pose measurement approaches such as those based on stereovision \cite{Li2020Super,Lu2020SuperDeep} or robot-state information \cite{Oneill2018Gripper,kong2018grip,stephens2019gripconditions} will need to be further developed to improve real-time accuracy. Even with the limitation in pose measurement accuracy, the dual jaw sensor meets the human perception-based performance specifications and is a promising tool for enabling RMIS research that requires force information for bimanual tasks. The sensor designs have been made available at \url{https://github.com/enhanced-telerobotics/RMIS_force_sensor/} to allow researchers to manufacture and modify the sensor for use in applications such as providing haptic feedback, performing robotic force control, and collecting bimanual RMIS manipulation datasets that inform data-driven computational methods.


\section{Acknowledgements}

The authors would like to thank Mark Cutkosky, Thomas Daunizeau, Michael Raitor, and Guan Rong Tan for their guidance on electromechanical design and sensor calibration. The surgical robot CAD model is courtesy of Koray Okan.


\addtolength{\textheight}{-6.2cm}   









\bibliographystyle{IEEEtran}
\bibliography{IEEEabrv,biblio}

\end{document}